\documentclass[10pt,twocolumn,letterpaper]{article}

\usepackage{cvpr}
\usepackage{times}
\usepackage{epsfig}
\usepackage{graphicx}
\usepackage{amsmath}
\usepackage{amssymb}
\usepackage{multirow}
\usepackage{caption}
\newcommand{\tabincell}[2]{\begin{tabular}{@{}#1@{}}#2\end{tabular}}

\usepackage[breaklinks=true,bookmarks=false]{hyperref}

\cvprfinalcopy 

\def\cvprPaperID{5558} 

\begin{document}

\title{Learning Parallax Attention for Stereo Image Super-Resolution}

\author{Longguang Wang$^{1}$, Yingqian Wang$^{1}$, Zhengfa Liang$^{2}$, Zaiping Lin$^{1}$, Jungang Yang$^{1}$, Wei An$^{1}$, Yulan Guo$^{1*}$\\
	$^{1}$College of Electronic Science and Technology, National University of Defense Technology, China\\
	$^{2}$National Key Laboratory of Science and Technology on Blind Signal Processing, China\\
	{\tt\small \{wanglongguang15,yulan.guo\}@nudt.edu}}


\maketitle

\begin{abstract}
	Stereo image pairs can be used to improve the performance of super-resolution (SR) since additional information is provided from a second viewpoint. However, it is challenging to incorporate this information for SR since disparities between stereo images vary significantly. In this paper, we propose a parallax-attention stereo super-resolution network (PASSRnet) to integrate the information from a stereo image pair for SR. Specifically, we introduce a parallax-attention mechanism with a global receptive field along the epipolar line to handle different stereo images with large disparity variations. We also propose a new and the largest dataset for stereo image SR (namely, Flickr1024). Extensive experiments demonstrate that the parallax-attention mechanism can capture correspondence between stereo images to improve SR performance with a small computational and memory cost. Comparative results show that our PASSRnet achieves the state-of-the-art performance on the Middlebury, \textcolor{black}{KITTI 2012 and KITTI 2015} datasets.
	
\end{abstract}

\section{Introduction}
Super-resolution (SR) aims to reconstruct high-resolution (HR) images from their low-resolution (LR) counterparts. Recovering an HR image from a single shot is a long-standing problem \cite{2014-LearningaDeepConvolutionalNetworkforImageSuperResolution-Dong-184-199,2016-RealTimeSingleImageandVideoSuperResolutionUsinganEfficientSubPixelConvolutionalNeuralNetwork-Shi-1874-1883,2018-FastandAccurateSingleImageSuperResolutionViaInformationDistillationNetwork-Hui--}. Recently, dual cameras are becoming increasingly popular in mobile phones and autonomous vehicles. It is already demonstrated that subpixel shifts contained in LR stereo images can be used to improve SR performance \cite{2003-SuperResolutionImageReconstruction:aTechnicalOverview-Park-21-36}. However, since disparities between stereo images can vary significantly for different baselines, focal lengths, depths and resolutions, it is highly challenging to incorporate stereo correspondence for SR. 

\begin{figure}[bt]
	\centering
	\begin{minipage}[t]{0.9\linewidth}
		%
		\begin{minipage}[t]{0.339\linewidth}
			\includegraphics[width=0.95\linewidth]{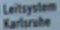}
			\setlength{\abovecaptionskip}{-11pt}
			\setlength{\belowcaptionskip}{-2pt}
			\caption*{Bicubic}
		\end{minipage}%
		\begin{minipage}[t]{0.339\linewidth}
			\includegraphics[width=0.95\linewidth]{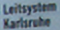}
			\setlength{\abovecaptionskip}{-11pt}
			\setlength{\belowcaptionskip}{-2pt}
			\caption*{SRCNN}
		\end{minipage}%
		\begin{minipage}[t]{0.339\linewidth}
			\includegraphics[width=0.95\linewidth]{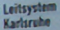}
			\setlength{\abovecaptionskip}{-11pt}
			\setlength{\belowcaptionskip}{-2pt}
			\caption*{LapSRN}
		\end{minipage}
		
		\begin{minipage}[t]{0.339\linewidth}
			\includegraphics[width=0.95\linewidth]{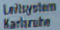}
			\setlength{\abovecaptionskip}{-11pt}
			\setlength{\belowcaptionskip}{0pt}
			\caption*{StereoSR}
		\end{minipage}%
		\begin{minipage}[t]{0.339\linewidth}
			\includegraphics[width=0.95\linewidth]{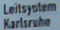}
			\setlength{\abovecaptionskip}{-11pt}
			\setlength{\belowcaptionskip}{0pt}
			\caption*{Ours}
		\end{minipage}%
		\begin{minipage}[t]{0.339\linewidth}
			\includegraphics[width=0.95\linewidth]{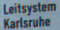}
			\setlength{\abovecaptionskip}{-11pt}
			\setlength{\belowcaptionskip}{0pt}
			\caption*{Groundtruth}
		\end{minipage}
	\end{minipage}%
	
	\caption{Visual results achieved by bicubic interpolation, SRCNN \cite{2014-LearningaDeepConvolutionalNetworkforImageSuperResolution-Dong-184-199}, LapSRN \cite{2017-DeepLaplacianPyramidNetworksforFastandAccurateSuperResolution-Lai-5835-5843}, StereoSR \cite{2018-EnhancingtheSpatialResolutionofStereoImagesUsingaParallaxPrior-Jeon--} and our network for $2\times$ SR. These results are achieved on ``test\_image\_002" of the KITTI 2015 dataset.}
	
	\label{fig1}
\end{figure}

Traditional multi-image SR methods \cite{2009-GeneralizingtheNonlocalMeanstoSuperResolutionReconstruction-Protter-36-51,2009-SuperResolutionwithoutExplicitSubpixelMotionEstimation-Takeda-1958-1975} use patch recurrence across images to obtain correspondence. However, these methods cannot \textcolor{black}{exploit} sub-pixel correspondence and their computational cost is high. Recent CNN-based frameworks \cite{2017-RealTimeVideoSuperResolutionwithSpatioTemporalNetworksandMotionCompensation-Caballero-2848-2857,2017-DetailRevealingDeepVideoSuperResolution-Tao-4482-4490,2018-LearningforVideoSuperResolutionthroughHROpticalFlowEstimation-LongguangWang--} incorporate optical flow estimation and SR in  unified networks to solve the video SR problem. However, these methods cannot be directly applied to stereo image SR since the disparity can be \textcolor{black}{much} larger than their receptive field.

Stereo matching has been investigated to obtain correspondence between \textcolor{black}{a stereo image pair} \cite{1982-ComputationalStereo-Barnard-553-572,2002-ATaxonomyandEvaluationofDenseTwoFrameStereoCorrespondenceAlgorithms-Scharstein--,2016-EfficientDeepLearningforStereoMatching-Luo-5695-5703}. Recent CNN-based methods \cite{2017-EndtoEndLearningofGeometryandContextforDeepStereoRegression-Kendall-66-75,2018-PyramidStereoMatchingNetwork-Chang--,2018-LearningDeepCorrespondencethroughPriorandPosteriorFeatureConstancy-Liang--,2018-LeftRightComparativeRecurrentModelforStereoMatching-Jie--} use 3D or 4D cost volumes in their \textcolor{black}{networks} to \textcolor{black}{model long-range dependency between stereo image pairs}. Intuitively, these CNN-based stereo matching methods can be	integrated with SR to provide accurate correspondence.  However, 4D cost volume based methods  \cite{2017-EndtoEndLearningofGeometryandContextforDeepStereoRegression-Kendall-66-75,2018-PyramidStereoMatchingNetwork-Chang--} suffer from a high computational and memory burden, which is unbearable for stereo image SR. Although the efficiency of 3D cost volume based methods  \cite{2018-LearningDeepCorrespondencethroughPriorandPosteriorFeatureConstancy-Liang--,2018-LeftRightComparativeRecurrentModelforStereoMatching-Jie--} is improved, these methods cannot handle stereo images with large disparity variations since a fixed maximum disparity is used to construct a cost volume.

Recently, Jeon \emph{et al.} proposed a stereo SR network (StereoSR) \cite{2018-EnhancingtheSpatialResolutionofStereoImagesUsingaParallaxPrior-Jeon--} to provide correspondence cues for SR using an image stack. Specifically, the image stack is obtained by concatenating the left image and the images generated by shifting the right image with different intervals. A direct mapping between parallax shifts and an HR image is then obtained. However, the flexibility of this method for different sensors and scenes is limited since the largest allowed disparity is fixed (\emph{i.e.}, 64 in \cite{2018-EnhancingtheSpatialResolutionofStereoImagesUsingaParallaxPrior-Jeon--}) in this algorithm.

In this paper, we propose a parallax-attention stereo SR network (PASSRnet) to incorporate stereo correspondence for the SR task. 
Given a stereo image pair, a residual atrous spatial pyramid pooling (ASPP) module is first used to generate multi-scale features.  Then, these features are fed to a parallax-attention module (PAM) to capture stereo correspondence. For each pixel in the left image, its feature similarities with all possible disparities in the right image are computed to generate an attention map. Consequently, our PAM can capture global correspondence while maintaining high flexibility. Afterwards, attention-driven feature aggregation is performed to update the features of the left image. Finally, these features are used to generate the SR result. Ablation study is performed on the KITTI 2015 dataset to test our PASSRnet. Comparative experiments are further conducted on the Middlebury, KITTI 2012 and KITTI 2015 datasets to demonstrate the superior performance of our network (as shown in Fig. \ref{fig1}).

Our main contributions can be summarized as follows: 
1) We propose a PASSRnet for SR by incorporating stereo correspondence;
2) We introduce a generic parallax-attention mechanism with a global receptive field along the epipolar line to handle different stereo images with large disparity variations. It is demonstrated that reliable correspondence can be efficiently generated by the parallax-attention mechanism for the improvement of SR performance;
3) We propose a new dataset, namely Flickr1024, \textcolor{black}{for the training of stereo image SR networks}. The Flickr1024 dataset consists of 1024 high-quality stereo image pairs and covers diverse scenes;
4) Our PASSRnet achieves the state-of-the-art performance as compared to recent single image SR and stereo image  SR methods.

\section{Related Work}
In this section, we briefly review several major works for SR and long-range dependency learning.
\subsection{Super-resolution}

\noindent 
\textbf{Single Image SR}
Since the seminal work of super-resolution convolutional neural network (SRCNN) \cite{2014-LearningaDeepConvolutionalNetworkforImageSuperResolution-Dong-184-199}, learning-based methods have dominated the research of single image SR. Kim \emph{et al.} \cite{2016-AccurateImageSuperResolutionUsingVeryDeepConvolutionalNetworks-Kim-1646-1654} proposed a very deep super-resolution network (VDSR) with 20 convolutional layers. Tai \emph{et al.} \cite{2017-ImageSuperResolutionViaDeepRecursiveResidualNetwork-Tai-2790-2798} developed a deep recursive residual network (DRRN) to control model parameters. Recently, Zhang \emph{et al.} \cite{2018-ResidualDenseNetworkforImageSuperResolution-Zhang--} proposed a residual dense network (RDN) to facilitate effective feature learning through a contiguous memory mechanism.

\noindent 
\textbf{Video SR}
\textcolor{black}{Liao \emph{et al.} \cite{2015-VideoSuperResolutionViaDeepDraftEnsembleLearning-Liao-531-539} introduced the first CNN for video SR.} They performed motion compensation to generate an ensemble of SR-drafts, and then employed a CNN to reconstruct  HR frames from the ensemble. Caballero \emph{et al.} \cite{2017-RealTimeVideoSuperResolutionwithSpatioTemporalNetworksandMotionCompensation-Caballero-2848-2857} proposed an end-to-end video SR framework by incorporating a motion compensation module with an SR module. Tao \emph{et al.} \cite{2017-DetailRevealingDeepVideoSuperResolution-Tao-4482-4490} \textcolor{black}{integrated an encoder-decoder network with LSTM} to fully use temporal correspondence. This architecture further facilitates the extraction of temporal context. Since correspondence between adjacent frames mainly exists within a local region, video SR methods focus on the \textcolor{black}{exploitation} of local dependency. Therefore, they cannot be directly applied to stereo image SR due to the non-local and long-range dependency in stereo images.

\noindent 
\textbf{Light-field Image SR}
Light-filed imaging can capture additional angular information of light at the cost of spatial resolution. To enhance spatial resolution, Yoon \emph{et al.} \cite{2015-LearningaDeepConvolutionalNetworkforLightFieldImageSuperResolution-Yoon-57-65} introduced \textcolor{black}{the first} light-field convolutional neural network (LFCNN). Yuan \emph{et al.} \cite{2018-LightFieldImageSuperresolutionUsingaCombinedDeepCNNBasedonEPI-Yuan-1359-1363a} proposed a CNN framework with a single image SR module and an epipolar plane image enhancement module. To model the correspondence between  images of adjacent sub-apertures, Wang \emph{et al.} \cite{2018-LFNet:aNovelBidirectionalRecurrentConvolutionalNeuralNetworkforLightFieldImageSuperResolution-Wang-4274-4286} developed a bidirectional recurrent CNN. Their network uses an implicit multi-scale feature fusion scheme to accumulate contextual information for SR. Note that, these methods 
are specifically proposed for light-field imaging with short baselines. Since stereo imaging usually has a much larger baseline than light-field imaging, these methods are unsuitable for stereo image SR.

\noindent 
\textbf{Stereo Image SR}
Bhavsar \emph{et al.} \cite{2010-ResolutionEnhancementinMultiImageStereo-Bhavsar-1721-1728} argued that image SR and HR depth estimation are intertwined \textcolor{black}{under stereo setting}. Therefore, they proposed an integrated approach to jointly estimate the SR image and HR disparity from LR stereo images. Recently, Jeon \emph{et al.} \cite{2018-EnhancingtheSpatialResolutionofStereoImagesUsingaParallaxPrior-Jeon--} proposed a StereoSR to employ parallax prior. Given a stereo image pair, the right image is shifted with different intervals and concatenated with the left image to generate a stereo tensor. The tensor is then fed to a plain CNN to generate the SR result by detecting similar patches within the disparity channel. However, StereoSR cannot handle different stereo images with large disparity variations since the number of shifted right images is fixed.

\begin{figure*}[ht]
	\centering
	\includegraphics[width=0.85\linewidth]{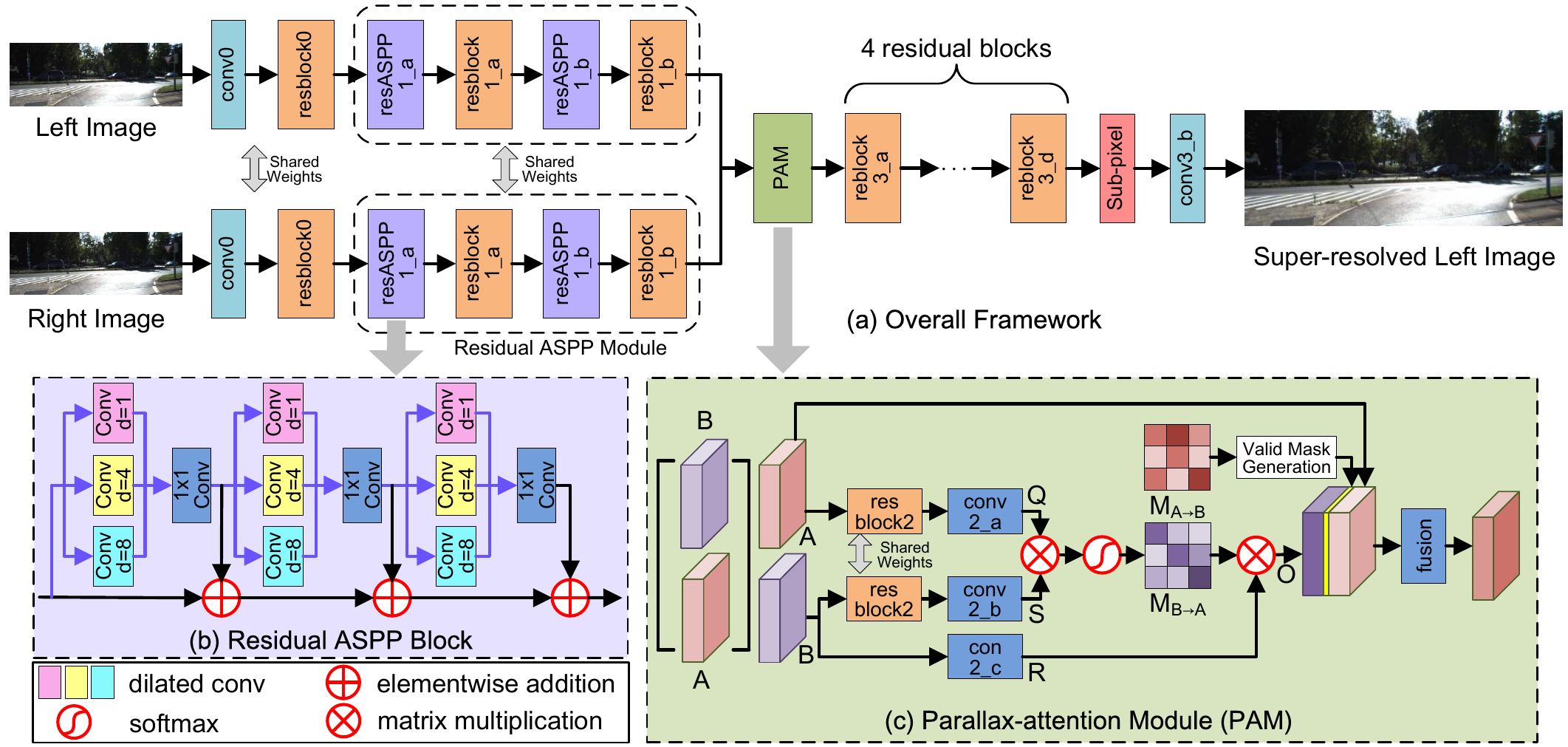}
	\caption{An overview of our PASSRnet.}
	\label{fig2}
\end{figure*}

\subsection{Long-range Dependency Learning}
To handle different stereo images with varying disparities for SR, long-range dependency in stereo images should be captured. In this section, we review two types of methods for long-range dependency learning.

\noindent
\textbf{\textcolor{black}{Cost Volume}}
Cost volume is widely applied in stereo matching \cite{2017-EndtoEndLearningofGeometryandContextforDeepStereoRegression-Kendall-66-75,2018-PyramidStereoMatchingNetwork-Chang--,2018-LearningDeepCorrespondencethroughPriorandPosteriorFeatureConstancy-Liang--} and optical flow estimation \cite{2017-PWCNet:CNNsforOpticalFlowUsingPyramidWarpingandCostVolume-Sun--,2017-AccurateOpticalFlowViaDirectCostVolumeProcessing-Xu-5807-5815}. \textcolor{black}{For stereo matching}, several methods \cite{2017-EndtoEndLearningofGeometryandContextforDeepStereoRegression-Kendall-66-75,2018-PyramidStereoMatchingNetwork-Chang--} use naive concatenation to construct 4D cost volumes. These methods concatenate left feature maps with their corresponding right feature maps across all disparities to obtain a 4D cost volume (\emph{i.e.}, height$\times${width}$\times${disparity}$\times${channel}). Then, 3D CNNs are usually used for matching cost learning. However, learning matching costs from 4D cost volumes suffers from a high computational and memory burden. To achieve higher efficiency, dot product is used to reduce feature dimension \cite{2018-LearningDeepCorrespondencethroughPriorandPosteriorFeatureConstancy-Liang--,2018-LeftRightComparativeRecurrentModelforStereoMatching-Jie--}, resulting in 3D cost volumes (\emph{i.e.}, height$\times${width}$\times${disparity}). However, due to the fixed maximum disparity in 3D cost volumes, these methods are unable to handle different \textcolor{black}{stereo image pairs} with large disparity variations.

\noindent
\textbf{\textcolor{black}{Self-attention Mechanisms}}
Attention mechanisms have been widely used to capture long-range dependency \cite{2015-DRAW:aRecurrentNeuralNetworkforImageGeneration-Gregor-1462-1471,2015-ShowAttendandTell:NeuralImageCaptionGenerationwithVisualAttention-Xu-2048-2057a}. For self-attention mechanisms \cite{2017-AttentionIsAllYouNeed-Vaswani-6000-6010,2018-SelfAttentionGenerativeAdversarialNetworks-Zhang--,2018-DualAttentionNetworkforSceneSegmentation-Fu--}, a weighted sum of all positions in spatial and/or temporal domain is calculated as the response at a position. Through matrix multiplication, self-attention mechanisms can capture the interaction between any two positions. Consequently, long-range dependency can be modeled with a small increase in computational and memory cost. Self-attention mechanisms have been successfully applied in image modeling \cite{2018-SelfAttentionGenerativeAdversarialNetworks-Zhang--} and semantic segmentation \cite{2018-DualAttentionNetworkforSceneSegmentation-Fu--}. Recent non-local networks \cite{2018-NonLocalNeuralNetworks-Wang--,2018-NonLocalRecurrentNetworkforImageRestoration-Liu--} share a similar idea and can be considered as a generalization of self-attention mechanisms. Note that, since self-attention mechanisms model dependency across the whole image, directly applying these mechanisms to stereo image SR involves unnecessary calculations.

Inspired by \textcolor{black}{self-attention mechanisms}, we develop a parallax-attention mechanism to model global dependency in stereo images. Compared to cost volumes, our parallax-attention mechanism is more flexible and efficient. Compared to self-attention mechanisms, our parallax-attention mechanism takes full use of epipolar constraints to reduce \textcolor{black}{search} space and improve efficiency. Moreover, the parallax-attention mechanism \textcolor{black}{enforces our network} to focus on the most similar feature rather than collecting all similar features for correspondence generation. It is demonstrated that the parallax-attention mechanism can generate reliable correspondence to improve SR performance (Section \ref{sec4.3.1}).

\section{Method}

Our PASSRnet takes a stereo image pair as input and super-resolves the left image. The architecture of our PASSRnet is shown in Fig. \ref{fig2} and Table \ref{tab1}.

\begin{table}[tp]
	\caption{The detailed architecture of our PASSRnet. LReLU represents leaky ReLU with a leakage factor of 0.1, dila stands for dilation rate, $\otimes$ denotes batch-wise matrix multiplication, and $s$ is the upscaling factor.}
	\label{tab1}
	\begin{center}
		\scriptsize
		\setlength{\tabcolsep}{0.5mm}{
			\begin{tabular}{|c|c|c|c|c|}
				\hline 
				{Name}  & {Setting} & {Input} & {Output} \tabularnewline
				\hline
				input & &$H\!\times\!{W}\!\times\!{3}$ & $H\!\times\!{W}\!\times\!{3}$ \tabularnewline
				\hline
				conv0 & \tabincell{c}{$3\!\times\!3$\\LReLU}  &$H\!\times\!{W}\!\times\!{3}$ &$H\!\times\!{W}\!\times\!{64}$ \tabularnewline
				\hline
				resblock0  & \Big[\tabincell{c}{$3\!\times\!3$\\$3\!\times\!3$}\Big] &$H\!\times\!{W}\!\times\!{64}$ &$H\!\times\!{W}\!\times\!{64}$ \tabularnewline
				\hline
				\multicolumn{4}{|c|}{\emph{Residual ASPP Module}}\tabularnewline
				\hline
				\tabincell{c}{resASPP\\1\_a} 
				&\Bigg[ \tabincell{c}{{\tabincell{c}{$3\!\times\!3$\\LReLU\\dila=1},
						\tabincell{c}{$3\!\times\!3$\\LReLU\\dila=4}, \tabincell{c}{$3\!\times\!3$\\LReLU\\dila=8}}\\$1\!\times\!1$} 
				\Bigg]$\!\times\!3$ 
				&$H\!\times\!{W}\!\times\!{64}$ &$H\!\times\!{W}\!\times\!{64}$\tabularnewline
				\hline
				\tabincell{c}{resblock\\1\_a}  & \Big[\tabincell{c}{$3\!\times\!3$\\$3\!\times\!3$}\Big] &$H\!\times\!{W}\!\times\!{64}$ &$H\!\times\!{W}\!\times\!{64}$ \tabularnewline
				\hline
				\tabincell{c}{resASPP\\1\_b} 
				& \Bigg[
				\tabincell{c}{{\tabincell{c}{$3\!\times\!3$\\LReLU\\dila=1},
						\tabincell{c}{$3\!\times\!3$\\LReLU\\dila=4}, \tabincell{c}{$3\!\times\!3$\\LReLU\\dila=8}}\\$1\!\times\!1$} 
				\Bigg]$\!\times\!3$  
				&$H\!\times\!{W}\!\times\!{64}$ &$H\!\times\!{W}\!\times\!{64}$
				\tabularnewline
				\hline
				\tabincell{c}{resblock\\1\_b}  & \Big[\tabincell{c}{$3\!\times\!3$\\$3\!\times\!3$}\Big] &$H\!\times\!{W}\!\times\!{64}$ &$H\!\times\!{W}\!\times\!{64}$ \tabularnewline
				\hline
				\multicolumn{4}{|c|}{\emph{Parallax-Attention Module}}\tabularnewline
				\hline
				resblock2  & \Big[\tabincell{c}{$3\!\times\!3$\\$3\!\times\!3$}\Big] &$H\!\times\!{W}\!\times\!{64}$ &$H\!\times\!{W}\!\times\!{64}$ \tabularnewline
				\hline
				conv2\_a & \tabincell{c}{$1\!\times\!1$} &$H\!\times\!{W}\!\times\!{64}$ &$H\!\times\!{W}\!\times\!{64}$ \tabularnewline
				\hline
				conv2\_b & \tabincell{c}{$1\times1$, reshape} &$H\!\times\!{W}\!\times\!{64}$ &$H\!\times\!{64}\!\times\!{W}$ \tabularnewline
				\hline
				conv2\_c & \tabincell{c}{$1\!\times\!1$} &$H\!\times\!{W}\!\times\!{64}$ &$H\!\times\!{W}\!\times\!{64}$ \tabularnewline
				\hline
				att\_map & {conv2\_a} $\otimes$ {conv2\_b}  & \tabincell{c}{$H\!\times\!{W}\!\times\!{64}$\\$H\!\times\!{64}\!\times\!{W}$} &$H\!\times\!{W}\!\times\!{W}$ \tabularnewline
				\hline
				mult & {att\_map} $\otimes$ {conv2\_c}  & \tabincell{c}{$H\!\times\!{W}\!\times\!{W}$\\$H\!\times\!{W}\!\times\!{64}$} &$H\!\times\!{W}\!\times\!{64}$ \tabularnewline
				\hline
				fusion & \tabincell{c}{$1\!\times\!1$} &$H\!\times\!{W}\!\times\!{129}$ &$H\!\times\!{W}\!\times\!{64}$ \tabularnewline
				\hline
				\multicolumn{4}{|c|}{\emph{CNN}}\tabularnewline
				\hline
				\tabincell{c}{resblock3\\$\!\times\!4$}  & \Big[\tabincell{c}{$3\!\times\!3$\\$3\!\times\!3$}\Big] &$H\!\times\!{W}\!\times\!{64}$ &$H\!\times\!{W}\!\times\!{64}$ \tabularnewline
				\hline
				sub-pixel & $1\!\times\!1$, pixel shuffle &$H\!\times\!{W}\!\times\!{64}$ &$sH\!\times\!{sW}\!\times\!{64}$ \tabularnewline
				\hline
				conv3\_b & $3\!\times\!3$ &$sH\!\times\!{sW}\!\times\!{64}$ &$sH\!\times\!{sW}\!\times\!{3}$ \tabularnewline
				\hline
		\end{tabular}}
	\end{center}
\end{table}

\subsection{Residual Atrous Spatial Pyramid Pooling (ASPP) Module}
Feature representation with rich context information is important for correspondence estimation \cite{2018-PyramidStereoMatchingNetwork-Chang--}. Therefore, both large receptive filed and multi-scale feature learning are required to obtain a discriminative representation. To this end, we propose a residual ASPP module to enlarge the receptive field and extract hierarchical features with dense pixel sampling rate and scales.

As shown in Fig. \ref{fig2} (a), our residual ASPP module is constructed by alternately cascading a residual ASPP block with a residual block. Input features are first fed to a residual ASPP block to generate multi-scale features. These resulting features are then sent to a residual block for feature fusion. This structure is repeated twice to produce final features. Within each residual ASPP block (as shown in Fig. \ref{fig2} (b)), we first combine three dilated convolutions (with dilation rates of 1, 4, 8) to form an ASPP group, and then cascade three ASPP groups in a residual manner. Our residual ASPP block not only enlarges the receptive field, but also enriches the diversity of convolutions, resulting in an ensemble of convolutions with different receptive regions and dilation rates. The highly discriminative feature learned by our residual ASPP module is beneficial to the overall SR performance, as demonstrated in Sec. \ref{sec4.3.1}.

\subsection{Parallax-attention Module (PAM)}
\label{sec3.2}
Inspired by \textcolor{black}{self-attention} mechanisms \cite{2018-SelfAttentionGenerativeAdversarialNetworks-Zhang--,2018-DualAttentionNetworkforSceneSegmentation-Fu--}, we develop PAM to capture global correspondence in stereo images. Our PAM efficiently integrates the information from \textcolor{black}{a stereo image pair}.

\noindent
\textbf{Parallax-attention Mechanism}
The architecture of our PAM is illustrated in Fig. \ref{fig2} (c). Given two feature maps ${\rm\textbf{A}},{\rm\textbf{B}}\!\in\!\mathbb{R}^{{H}\times{W}\times{C}}$, they are fed to a transition residual block to generate $\rm\textbf{A}_{0}$ and $\rm\textbf{B}_{0}$. Then, $\rm\textbf{A}_{0}$ is fed to a $1\times1$ convolution layer to produce a query feature map ${\rm\textbf{Q}}\!\in\!\mathbb{R}^{{H}\times{W}\times{C}}$. 
Meanwhile, ${\rm\textbf{B}}_{0}$ is fed to another $1\times1$ convolution layer to produce  ${\rm\textbf{S}}\!\in\!\mathbb{R}^{{H}\times{W}\times{C}}$, which is then reshaped to $\mathbb{R}^{{H}\times{C}\times{W}}$. Batch-wise matrix multiplication is then performed between $\rm\textbf{Q}$ and $\rm\textbf{S}$ and a softmax layer is applied, resulting in a parallax attention map ${\rm\textbf{M}}_{{\rm\textbf{B}}\rightarrow{\rm\textbf{A}}}\!\in\!\mathbb{R}^{{H}\times{W}\times{W}}$. For more details, please refer to the supplemental material. Next, $\rm\textbf{B}$ is fed to a $1\times1$ convolution to generate ${\rm\textbf{R}}\!\in\!\mathbb{R}^{{H}\times{W}\times{C}}$, which is further multiplied \textcolor{black}{by} ${\rm\textbf{M}}_{\rm\textbf{B}\rightarrow{\rm\textbf{A}}}$ to produce features ${\rm\textbf{O}}\!\in\!\mathbb{R}^{{H}\times{W}\times{C}}$. As a weighted sum of features at all possible disparities, $\rm\textbf{O}$ is  then integrated with its corresponding local features $\rm\textbf{A}$. Since PAM can gradually focus on the features at \textcolor{black}{accurate} disparities using feature similarities, correspondence can then be captured. Note that, once ${\rm\textbf{M}}_{{\rm\textbf{B}}\rightarrow{\rm\textbf{A}}}$ is ready, $\rm\textbf{A}$ and $\rm\textbf{B}$ are exchanged to produce  ${\rm\textbf{M}}_{{\rm\textbf{A}}\rightarrow{\rm\textbf{B}}}$ for valid mask generation (as described below). Finally, stacked features and a valid mask  are fed to a $1\times1$ convolution layer for feature fusion.

\begin{figure}[tp]
	\centering
	\includegraphics[width=0.95\linewidth]{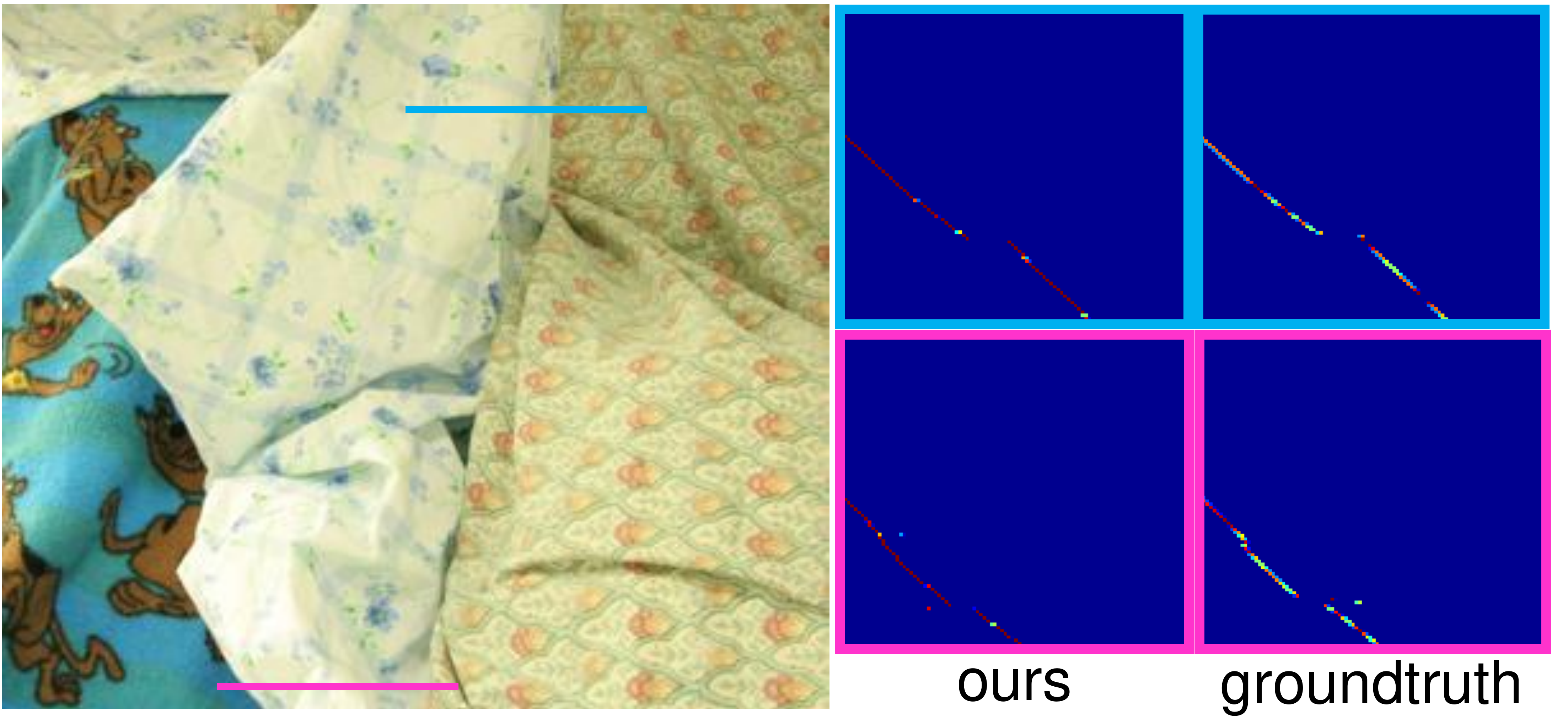}
	\caption{Visual comparison between parallax-attention maps ${\rm\textbf{M}}_{\emph{right}\rightarrow\emph{left}}$ generated by our PAM and the groundtruth. These attention maps ($100\!\times\!100$) correspond to the regions ($1\!\times\!100$) marked by blue and pink strokes in the left image.}
	\label{fig5}
\end{figure}

Different from self-attention mechanisms \cite{2018-SelfAttentionGenerativeAdversarialNetworks-Zhang--,2018-DualAttentionNetworkforSceneSegmentation-Fu--}, our parallax-attention mechanism \textcolor{black}{enforces our network} to focus on the most similar feature along the epipolar line rather than collecting all similar features, resulting in sparse attention maps. A comparison between parallax-attention maps generated by our PAM and the groundtruth is shown in Fig. \ref{fig5}. 
Note that, ${\rm\textbf{M}}_{\emph{right}\rightarrow\emph{left}}(i,j,k)$ represents the contribution of position $(i,k)$ in the right image to position $(i,j)$ in the left image. Consequently, the patterns in an attention map can reflect the correspondence between stereo pairs and also encode disparity information.
For more details, please refer to the supplemental material. It can be observed that our PAM produces patterns similar to the groundtruth, which indicates that reliable stereo correspondence can be captured by our PAM. It should be noted that our PASSRnet can be considered as a multi-task network to learn both stereo correspondence and SR. \textcolor{black}{However, using shared features for different tasks usually suffers from training conflict \cite{2018-MultiTaskLearningAsMultiObjectiveOptimization-Sener--}. Therefore, a transition block is introduced in our PAM to alleviate this problem.} The effectiveness of the transition block is demonstrated in Sec. \ref{sec4.3.1}.

\noindent
\textbf{Left-right Consistency \& Cycle Consistency}
Given deep features extracted from an LR stereo image pair ($\rm\textbf{I}_\emph{left}^{L}$ and $\rm\textbf{I}_\emph{right}^{L}$), two parallax-attention maps ($\rm\textbf{M}_{\emph{left}\rightarrow\emph{right}}$ and $\rm\textbf{M}_{\emph{right}\rightarrow\emph{left}}$) can be generated by PAM. Ideally, the following left-right consistency can be obtained if our PAM captures accurate correspondence:
\begin{equation}
	\label{eq2}
	\left\{
	\begin{aligned}
		\rm\textbf{I}_\emph{left}^{L}&\!=\!\rm\textbf{M}_{\emph{right}\rightarrow\emph{left}}\otimes\rm\textbf{I}_\emph{right}^{L}\\
		\rm\textbf{I}_\emph{right}^{L}&\!=\!\rm\textbf{M}_{\emph{left}\rightarrow\emph{right}}\otimes\rm\textbf{I}_\emph{left}^{L}
	\end{aligned}
	\right.,
\end{equation}
where $\otimes$ denotes batch-wise matrix multiplication. Based on Eq. (\ref{eq2}), we can further derive a cycle consistency:
\begin{equation}
	\label{eq3}
	\left\{
	\begin{aligned}
		\rm\textbf{I}_\emph{left}^{L}&\!=\!\rm\textbf{M}_{\emph{left}\rightarrow\emph{right}\rightarrow\emph{left}}\otimes\rm\textbf{I}_\emph{left}^{L}\\
		\rm\textbf{I}_\emph{right}^{L}&\!=\!\rm\textbf{M}_{\emph{right}\rightarrow\emph{left}\rightarrow\emph{right}}\otimes\rm\textbf{I}_\emph{right}^{L}
	\end{aligned}
	\right.,
\end{equation}
where the cycle-attention maps $\rm\textbf{M}_{\emph{left}\rightarrow\emph{right}\rightarrow\emph{left}}$ and $\rm\textbf{M}_{\emph{right}\rightarrow\emph{left}\rightarrow\emph{right}}$ can be calculated as:
\begin{equation}
	\left\{
	\begin{aligned}
		\rm\textbf{M}_{\emph{left}\rightarrow\emph{right}\rightarrow\emph{left}}&\!=\!\rm\textbf{M}_{\emph{right}\rightarrow\emph{left}}\otimes\rm\textbf{M}_{\emph{left}\rightarrow\emph{right}}\\
		\rm\textbf{M}_{\emph{right}\rightarrow\emph{left}\rightarrow\emph{right}}&\!=
		\!\rm\textbf{M}_{\emph{left}\rightarrow\emph{right}}\otimes\rm\textbf{M}_{\emph{right}\rightarrow\emph{left}}
	\end{aligned}
	\right..
\end{equation}
Here, we introduce left-right consistency and cycle consistency to regularize  the training of our PAM for the generation of reliable and consistent correspondence.

\begin{figure}[bt]
	\centering
	\includegraphics[width=0.95\linewidth]{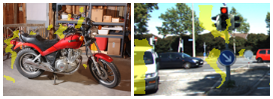}
	\caption{Visualization of valid masks. Two left images and their occluded regions (\emph{i.e.}, yellow regions) are illustrated.}
	\label{fig6}
\end{figure}

\noindent
\textbf{Valid Masks}
Since left-right consistency and cycle consistency do not hold for occluded regions, we use an occlusion detection method to generate valid masks. We only enforce consistency on valid regions. In the parallax-attention map generated by our PAM (\emph{e.g.}, $\rm\textbf{M}_{\emph{left}\rightarrow\emph{right}}$), it is observed that pixels in occluded regions are usually assigned with small weights. Therefore, a valid mask ${\rm\textbf{V}}_{\emph{left}\rightarrow\emph{right}}\!\in\!\mathbb{R}^{{H}\times{W}}$ can be obtained by:
\begin{equation}
	{\rm\textbf{V}}_{\emph{left}\rightarrow\emph{right}}(i,j)\!=\!
	\!\left\{\!
	\begin{array}{lr}
		1,&if  \sum_{{k}\!\in\![1,\,{W}]}{{\rm\textbf{M}}_{\emph{left}\rightarrow\emph{right}}(i,k,j)}>\textcolor{black}{\tau}, \\
		0,&otherwise,  
	\end{array}
	\right.
\end{equation}
where $\textcolor{black}{\tau}$ is a threshold (empirically set to 0.1 in this paper) and $W$ is the width of stereo images. Two examples of valid masks are shown in Fig. \ref{fig6}. According to the parallax-attention mechanism, ${\rm\textbf{M}}_{\emph{left}\rightarrow\emph{right}}(i,k,j)$ represents the contribution of position $(i,j)$ in the left image to position $(i,k)$ in the right image. Since occluded pixels in the left image cannot find their correspondences in the right image, their values ${\rm\textbf{V}}_{\emph{left}\rightarrow\emph{right}}(i,j)$ are usually low. Thus, we consider these pixels as occluded ones. In practice, we use several morphological operations to handle isolated pixels and holes in valid masks. Note that, occluded regions in the left image cannot obtain additional information from the right image. Therefore, valid mask $\rm\textbf{V}_{\emph{left}\rightarrow\emph{right}}$ is further used to guide feature fusion, as shown in Fig. \ref{fig2} (c).

\subsection{Losses}
\label{sec3.3}
We design four losses for the training of our PASSRnet. Other than an SR loss, we introduce three additional losses, including photometric loss, smoothness loss and cycle loss, to \textcolor{black}{help the network to fully exploit} the correspondence between stereo images. The overall loss function is formulated as:
\begin{equation}
	\mathcal{L}=\mathcal{L}_{\mathrm{SR}}+\lambda(\mathcal{L}_{\mathrm{photometric}}+\mathcal{L}_{\mathrm{smooth}}+\mathcal{L}_{\mathrm{cycle}}),
\end{equation}
where $\lambda$ is empirically set to 0.005. The performance of our network with different losses will be analyzed in Sec. \ref{sec4.3.2}.

\noindent
\textbf{SR Loss}
The mean square error (MSE) loss ifs used as the SR loss:
\begin{equation}
	\mathcal{L_{\mathrm{SR}}}=\left\Vert \rm\textbf{I}_\emph{left}^{SR}-\rm\textbf{I}_\emph{left}^{H}\right\Vert _{2}^{2},
\end{equation}
where $\rm\textbf{I}_\emph{left}^{SR}$ and $\rm\textbf{I}_\emph{left}^{H}$ represent the SR result and HR groundtruth of the left image, respectively.

\noindent
\textbf{Photometric Loss}
Since collecting a large stereo dataset with densely labeled groundtruth disparities is highly challenging, we train our PAM in an unsupervised manner. Note that, if the groundtruth disparities are available, we can generate the groundtruth attention maps accordingly (see the supplemental material for more details) and train our PAM in a supervised manner. Following \cite{2017-UnsupervisedMonocularDepthEstimationwithLeftRightConsistency-Godard-6602-6611}, we introduce a photometric loss using the mean absolute error (MAE) loss. Note that, since the left-right consistency defined in Eq. (\ref{eq2}) only holds in non-occluded regions, we introduce a photometric loss as:
\begin{equation}
	\begin{aligned}
		\mathcal{L_{\mathrm{photometric}}}\!=\!&\sum_{p\!\in\!{{\rm\textbf{V}}_{\emph{left}\rightarrow\emph{right}}}}\!\left\Vert {\rm\textbf{I}}_\emph{left}^{\rm L}(p)\!-\!(\rm\textbf{M}_{\emph{right}\rightarrow\emph{left}}\otimes{\rm\textbf{I}}_\emph{right}^{L})(p)\right\Vert_{1}\\\!+\!
		&\sum_{p\!\in\!{\rm\textbf{V}_{\emph{right}\rightarrow\emph{left}}}}\!\left\Vert \rm\textbf{I}_\emph{right}^{L}(p)\!-\!(\rm\textbf{M}_{\emph{left}\rightarrow\emph{right}}\otimes\rm\textbf{I}_\emph{left}^{L})(p)\right\Vert_{1},
	\end{aligned}
\end{equation}
where $p$ represents a pixel with a valid mask value.

\noindent
\textbf{Smoothness Loss}
To generate accurate and consistent attention in textureless regions, a smoothness loss is defined on the attention maps $\rm\textbf{M}_{\emph{left}\rightarrow\emph{right}}$ and $\rm\textbf{M}_{\emph{right}\rightarrow\emph{left}}$:
\begin{equation}
	\begin{aligned}
		\mathcal{L_{\mathrm{smooth}}}\!=\!\sum_{\rm\textbf{M}}\sum_{i,j,k}(&\left\Vert {\rm\textbf{M}}(i,j,k)\!-\!{\rm\textbf{M}}(i\!+\!1,j,k)\right\Vert_{1}\!+\!\\
		&\left\Vert {\rm\textbf{M}}(i,j,k)-\!{\rm\textbf{M}}(i,j\!+\!1,k\!+\!1)\right\Vert_{1}),
	\end{aligned}
	\label{eq8}
\end{equation}
where $\rm\textbf{M}\!\in\!\left\{\rm\textbf{M}_{\emph{left}\rightarrow\emph{right}},\rm\textbf{M}_{\emph{right}\rightarrow\emph{left}}\right\}$. The first and second terms in Eq. (\ref{eq8}) are used to achieve vertical and horizontal attention consistency, respectively. 

\noindent
\textbf{Cycle Loss}
In addition to photometric loss and smoothness loss, we further introduce a cycle loss to achieve cycle consistency. Since $\rm\textbf{M}_{\emph{left}\rightarrow\emph{right}\rightarrow\emph{left}}$ and $\rm\textbf{M}_{\emph{right}\rightarrow\emph{left}\rightarrow\emph{right}}$ in Eq. (\ref{eq3}) can be considered as identity matrices, we design a cycle loss as:
\begin{equation}
	\begin{aligned}
		\mathcal{L_{\mathrm{cycle}}}=&\sum_{p\!\in\!{{\rm\textbf{V}}_{\emph{left}\rightarrow\emph{right}}}}\left\Vert {\rm\textbf{M}}_{{\emph{left}\rightarrow\emph{right}\rightarrow\emph{left}}}(p)-I(p)\right\Vert_{1}+\\
		&\sum_{p\!\in\!{{\rm\textbf{V}}_{\emph{right}\rightarrow\emph{left}}}}\left\Vert{\rm\textbf{M}}_{{\emph{right}\rightarrow\emph{left}\rightarrow\emph{right}}}(p)-I(p)\right\Vert_{1},
	\end{aligned}
\end{equation}
where  $I\!\in\!\mathbb{R}^{{H}\times{W}\times{W}}$ is a stack of \textcolor{black}{$H$} identity matrices.

\begin{table*}[ht]
	\caption{Comparative results achieved on the KITTI 2015 dataset by PASSRnet with different settings for $4\times$ SR.}
	\label{tab2}
	\begin{center}
		\footnotesize
		\setlength{\tabcolsep}{2mm}{
			\begin{tabular}{|l|c|c|c|c|c|}
				\hline 
				Model & Input & PSNR & SSIM & Params. & Time
				\tabularnewline
				\hline
				PASSRnet with single input & Left & 25.27 & 0.770  & \textbf{1.32M} & \textbf{114ms} 
				\tabularnewline
				PASSRnet with replicated inputs & Left-Left & 25.29 & 0.771 & 1.42M & 176ms 
				\tabularnewline
				\hline
				PASSRnet without residual manner & Left-Right & 25.40 & 0.774  & 1.42M&176ms
				\tabularnewline
				PASSRnet without atrous convolution  & Left-Right & 25.38  & 0.773   &1.42M & 176ms 
				\tabularnewline
				\hline
				PASSRnet without PAM & Left-Right & 25.28  & 0.771   & \textbf{1.32M} & 135ms
				\tabularnewline
				PASSRnet without transition residual block & Left-Right & 25.36 & 0.773 & 1.34M & 160ms 
				\tabularnewline
				\hline				
				PASSRnet  & Left-Right & \textbf{25.43} & \textbf{0.776} & 1.42M & 176ms 
				\tabularnewline
				\hline
		\end{tabular}}
	\end{center}
\end{table*}

\section{Experimental Results}
In this section, we first introduce the datasets and implementation details, and then conduct ablation experiments to test our network. We further compare our network to recent single image SR and stereo image SR methods.

\subsection{\textcolor{black}{Datasets}}
For training, we followed \cite{2018-EnhancingtheSpatialResolutionofStereoImagesUsingaParallaxPrior-Jeon--} and downsampled 60 Middlebury \cite{2014-HighResolutionStereoDatasetswithSubpixelAccurateGroundTruth-Scharstein-31-42} images by a factor of 2 to generate HR images. We further collected 1024  stereo images from Flickr to construct a new Flickr1024 dataset. This dataset was used as the augmented training data for our PASSRnet. Please see the supplemental material for more details about the Flickr1024 dataset. For test, we used 5 images from the Middlebury dataset, 20 images from the KITTI 2012 dataset \cite{2012-AreWeReadyforAutonomousDriving?theKITTIVisionBenchmarkSuite-Geiger-3354-3361} and 20 images from the KITTI 2015 dataset \cite{2015-ObjectSceneFlowforAutonomousVehicles-Menze-3061-3070} as benchmark \textcolor{black}{datasets}. We further collected 10 close-shot stereo images (with disparities larger than 200) from Flickr to test the flexibility of our network to large disparity variations. For validation, we selected another 20 images from
the KITTI 2012 dataset.

\subsection{Implementation Details}
During the training phase, we first downsampled HR images using bicubic interpolation to generate LR images, and then cropped $30\times90$ patches with a stride of 20 from these LR images. Meanwhile, their corresponding patches in HR images were also cropped. The horizontal patch size was increased to 90 to cover most disparities ($\sim$96\%) in our training dataset. These patches were randomly flipped horizontally and vertically for data augmentation. Note that, rotation was not performed to maintain epipolar constraints. We used peak signal-to-noise ratio (PSNR) and structural similarity index (SSIM) to test SR performance. Similar to \cite{2018-EnhancingtheSpatialResolutionofStereoImagesUsingaParallaxPrior-Jeon--}, we cropped borders to achieve fair comparison.

Our PASSRnet was implemented in Pytorch on a PC with an Nvidia GTX 1080Ti GPU. All models were optimized using the Adam method \cite{2015-Adam:aMethodforStochasticOptimization-Kingma--} with $\beta_{1}=0.9$, $\beta_{2}=0.999$ and a batch size of 32. The initial learning rate was set to $2\!\times\!10^{-4}$ and reduced to half after every 30 epochs. The training was stopped after 80 epochs since more epochs do not provide further consistent improvement.

\subsection{Ablation Study}
In this section, we present ablation experiments to justify our design choices, including the network architecture and the losses.

\subsubsection{Network Architecture}
\label{sec4.3.1}
\noindent
\textbf{Single Input vs. Stereo Input}
Compared to single images, stereo image pairs provide additional information observed from a different viewpoint. To demonstrate the effectiveness of stereo information for SR performance improvement, we removed PAM from our PASSRnet and retrained the network with single images (\emph{i.e.}, the left images). For comparison, we also used pairs of replicated left images as the input to the original PASSRnet. Results achieved on the KITTI 2015 dataset are listed in Table \ref{tab2}. 

Compared to the original PASSRnet, the network trained with single images suffers a decrease of 0.16 dB (25.43 to 25.27) in PSNR. Further, if pairs of replicated left images are fed to the original PASSRnet, the PSNR value is decreased to 25.29 dB. Without extra information introduced by stereo images, our PASSRnet with replicated images achieves comparable performance to the network trained with single images. This clearly demonstrates that stereo images can be used to improve the performance of PASSRnet.

\noindent
\textbf{Residual ASPP Module}
Residual ASPP module is used in our network to extract multi-scale features. To demonstrate the effectiveness of residual ASPP, two variants were introduced. First, to test the effectiveness of residual connections, we removed them to obtain a cascading ASPP module. Then, to test the effectiveness of atrous convolutions, we replaced them with ordinary convolutions. 

From the comparative results shown in Table \ref{tab2}, we can see that SR performance benefits from both residual connections and atrous convolutions. If residual connections are removed, the PSNR value is decreased from 25.43 dB to 25.40 dB. That is because, residual connections enable our residual ASPP module to extract features at more scales, resulting in more robust feature representations. Furthermore, if atrous convolutions are replaced by ordinary ones, the PSNR value is decreased from 25.43 dB to 25.38 dB. That is because, large receptive field of atrous convolutions facilitates our PASSRnet to employ context information in a large area. Therefore, more accurate correspondence can be obtained to improve SR performance.

\noindent
\textbf{Parallax-attention Module}
PAM is introduced to integrate the information from stereo images. To demonstrate its effectiveness, we introduced a variant by removing PAM and directly stacking the output features of the residual ASPP module. It can be observed from Table \ref{tab2} that the PSNR value is decreased from 25.43 dB to 25.28 dB if PAM is removed. That is because, long spatial distance between local features in the left image and their dependency in the right image hinders plain CNNs to integrate these features effectively.

\noindent
\textbf{Transition Block in PAM}
Transition block in PAM is introduced to alleviate the training conflict in shared layers. To demonstrate the effectiveness of transition block, we removed it from our PAM and retrained the network.
It can be observed from Table \ref{tab2} that the PSNR value is decreased  from 25.43 dB to 25.36 dB if the transition block is removed. That is because, the  transition block enhances task-specific feature learning in our PAM and alleviates training conflict in shared layers. Therefore, more representative features can be learned in shared layers.

\noindent
\textbf{PAM vs. Cost Volume}
Cost volume and 3D convolutions are commonly used to obtain stereo correspondence \cite{2017-EndtoEndLearningofGeometryandContextforDeepStereoRegression-Kendall-66-75,2018-PyramidStereoMatchingNetwork-Chang--}. To demonstrate the efficiency of our PAM in stereo correspondence generation, we replaced PAM with a 4D cost volume and two 3D convolutional layers ($3\!\times\!3\!\times3$). It can be observed from Table \ref{tab3} that our PAM has less than half of the parameters in the cost volume formation. Moreover, our PAM achieves superior computational efficiency, with FLOPs being reduced by over 150 times. \textcolor{black}{ With PAM, our PASSRnet achieves better SR performance (\emph{i.e.}, PSNR value is increased from 25.23 dB to 25.43 dB) and efficiency (\emph{i.e.}, running time is decreased by 1.5 times).} That is because, two 3D convolutional layers are insufficient to capture long-range correspondence within the cost volume. However, adding more layers will lead to a significant increase of computational cost.

\begin{table}[bt]
	\caption{Comparison between our PAM and the cost volume formation for $4\times$ SR. FLOPs are calculated on $128\!\times\!128\!\times\!64$ input features, while Time/PSNR/SSIM values are achieved on the KITTI 2015 dataset.}
	\label{tab3}
	\begin{center}
		\footnotesize
		\setlength{\tabcolsep}{1mm}{
			\begin{tabular}{|c|c|c|c|c|c|}
				\hline 
				Model & Params. & FLOPs & Time & PSNR & SSIM
				\tabularnewline
				\hline
				PAM & 94K & $1\!\times$ & $1\!\times$ & 25.43 & 0.776 \tabularnewline
				\hline
				Cost Volume  & 221K & $151\!\times$ & $\!1.5\times$ & 25.23 & 0.768 
				\tabularnewline
				\hline
		\end{tabular}}
	\end{center}
\end{table}

\subsubsection{Losses}
\label{sec4.3.2}
To test the effectiveness of our losses, we retrained PASSRnet using different losses.

It can be observed from Table \ref{tab4} that the PSNR value of our PASSRnet is decreased from 25.43 to 25.35 if PASSRnet is trained with only SR loss. \textcolor{black}{That is because, with only this loss, our PAM learns to collect all similar features along the epipolar line and cannot focus on the most similar feature to provide accurate correspondence.} Further, the performance is gradually improved if photometric loss, smoothness loss and cycle loss are added. That is because, these losses encourage our PAM to generate reliable and consistent correspondence. Overall, our PASSRnet achieves the best performance (\emph{i.e.}, PSNR=25.43 dB and SSIM=0.776) when it is trained with all these losses.

\begin{table}[tp]
	\caption{Comparative results achieved on KITTI 2015 by our PASSRnet trained with different losses for $4\times$ SR.}
	\label{tab4}
	\begin{center}
		\footnotesize
		\setlength{\tabcolsep}{0.8mm}{
			\begin{tabular}{|c|cccc|c|c|}
				\hline 
				Model & $\mathcal{L}_{SR}$ & $\mathcal{L}_{photometric}$ & $\mathcal{L}_{smooth}$ & $\mathcal{L}_{cycle}$ & PSNR & SSIM 
				\tabularnewline
				\hline
				PASSRnet & \checkmark   &   &   &  & 25.35 & 0.771  \tabularnewline
				PASSRnet & \checkmark & \checkmark &   &  & 25.38 & 0.773 
				\tabularnewline
				PASSRnet & \checkmark & \checkmark & \checkmark &  & 25.40 & 0.774 
				\tabularnewline
				PASSRnet & \checkmark & \checkmark & \checkmark & \checkmark& \textbf{25.43}& \textbf{0.776} \tabularnewline
				\hline
		\end{tabular}}
	\end{center}
\end{table}

\begin{table*}[ht]
	\caption{Comparative PSNR/SSIM values achieved on the Middlebury, KITTI 2012 and KITTI 2015 datasets. Results marked with * are directly copied from the corresponding paper. Note that, only 2$\times$ SR results of StereoSR are presented on the KITTI 2012 and KITTI 2015 datasets since a 4$\times$ SR model is unavailable.}
	\label{tab5}
	\begin{center}
		\footnotesize
		\setlength{\tabcolsep}{1.5mm}{
			\begin{tabular}{|c|c|ccccc||cc|}
				\hline 
				\multirow{2}{*}{Dataset} &  \multirow{2}{*}{Scale} & \multicolumn{5}{c||}{Single Image SR} & \multicolumn{2}{c|}{Stereo Image SR}
				\tabularnewline
				
				&  
				&SRCNN \cite{2014-LearningaDeepConvolutionalNetworkforImageSuperResolution-Dong-184-199} 
				&VDSR \cite{2016-AccurateImageSuperResolutionUsingVeryDeepConvolutionalNetworks-Kim-1646-1654}  
				&DRCN \cite{2016-DeeplyRecursiveConvolutionalNetworkforImageSuperResolution-Kim-1637-1645}
				&LapSRN \cite{2017-DeepLaplacianPyramidNetworksforFastandAccurateSuperResolution-Lai-5835-5843} 
				&DRRN  \cite{2017-ImageSuperResolutionViaDeepRecursiveResidualNetwork-Tai-2790-2798} 
				&StereoSR \cite{2018-EnhancingtheSpatialResolutionofStereoImagesUsingaParallaxPrior-Jeon--} 
				& Ours 
				\tabularnewline
				\hline
				\multirow{2}{*}{\tabincell{c}{Middlebury\\(5 images)}} & $\times$2 &32.05/0.935&32.66/0.941&32.82/0.941&32.75/0.940&32.91/0.945&33.05/0.955*&\textbf{34.05/0.960} 
				\tabularnewline
				& $\times$4 &27.46/0.843&27.89/0.853&27.93/0.856&27.98/0.861&27.93/0.855&26.80/0.850*&\textbf{28.63/0.871} 
				\tabularnewline
				\hline
				
				\multirow{2}{*}{\tabincell{c}{KITTI 2012\\(20 images)}} & $\times$2 &29.75/0.901&30.17/0.906&30.19/0.906&30.10/0.905&30.16/0.908&30.13/0.908&\textbf{30.65/0.916} 
				\tabularnewline
				& $\times$4 &25.53/0.764&25.93/0.778&25.92/0.777&25.96/0.779&25.94/0.773&-&\textbf{26.26/0.790} 
				\tabularnewline
				\hline
				\multirow{2}{*}{\tabincell{c}{KITTI 2015\\(20 images)}} & $\times$2 &28.77/0.901&28.99/0.904&29.04/0.904&28.97/0.903&29.00/0.906&29.09/0.909&\textbf{29.78/0.919} 
				\tabularnewline
				& $\times$4 &24.68/0.744&25.01/0.760&25.04/0.759&25.03/0.760&25.05/0.756&-&\textbf{25.43/0.776} 
				\tabularnewline
				\hline
		\end{tabular}}
	\end{center}
\end{table*}

\begin{figure*}[ht]
	\centering
	\includegraphics[width=0.98\linewidth]{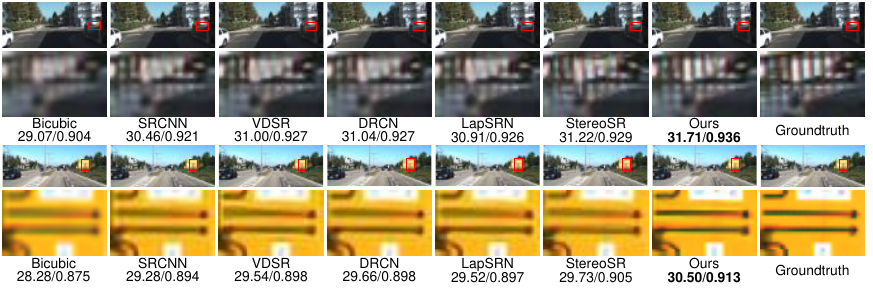}
	\caption{Visual comparison for $2\times$ SR. These results are achieved on ``test\_image\_013" of the KITTI 2012 dataset and ``test\_image\_019" of the KITTI 2015 dataset.}
	\label{fig7}
\end{figure*}


\textcolor{black}{
	\begin{table}[bt]
		\caption{Comparison between \textcolor{black}{our PASSRnet and StereoSR \cite{2018-EnhancingtheSpatialResolutionofStereoImagesUsingaParallaxPrior-Jeon--}} on stereo images with different resolutions for 2$\times$ SR.}
		\begin{center}
			\footnotesize
			\setlength{\tabcolsep}{1.2mm}{
				\begin{tabular}{|l|cc|cc|}
					\hline 
					\multirow{2}{*}{Resolution} &  \multicolumn{2}{c|}{StereoSR \cite{2018-EnhancingtheSpatialResolutionofStereoImagesUsingaParallaxPrior-Jeon--}} & \multicolumn{2}{c|}{Ours}
					\tabularnewline
					& PSNR & FLOPs & PSNR & FLOPs
					\tabularnewline
					\hline
					High ($500\times500$)  & 39.27 &\textcolor{black}{1}$\times$ & \textcolor{black}{\textbf{41.45}}($\uparrow2.18$) & \textcolor{black}{0.57}$\times$
					\tabularnewline
					\hline
					Middle ($100\times100$) & 34.21 & \textcolor{black}{1}$\times$ & \textcolor{black}{\textbf{35.04}}($\uparrow0.83$) & \textcolor{black}{0.58}$\times$
					\tabularnewline
					\hline
					Low ($20\times20$) & 29.48 & \textcolor{black}{1}$\times$ & \textcolor{black}{\textbf{29.88}}($\uparrow0.40$) & \textcolor{black}{0.36}$\times$
					\tabularnewline
					\hline
			\end{tabular}}
		\end{center}
		\label{tab6}
	\end{table}
}

\subsection{Comparison to State-of-the-arts}
We compared our PASSRnet to a number of CNN-based SR methods on \textcolor{black}{three benchmark datasets}. Recent single image SR methods under comparison include SRCNN \cite{2014-LearningaDeepConvolutionalNetworkforImageSuperResolution-Dong-184-199}, VDSR \cite{2016-AccurateImageSuperResolutionUsingVeryDeepConvolutionalNetworks-Kim-1646-1654}, DRCN \cite{2016-DeeplyRecursiveConvolutionalNetworkforImageSuperResolution-Kim-1637-1645}, LapSRN \cite{2017-DeepLaplacianPyramidNetworksforFastandAccurateSuperResolution-Lai-5835-5843} and DRRN \cite{2017-ImageSuperResolutionViaDeepRecursiveResidualNetwork-Tai-2790-2798}. We also compared our PASSRnet to the latest stereo image SR method StereoSR \cite{2018-EnhancingtheSpatialResolutionofStereoImagesUsingaParallaxPrior-Jeon--}. The codes provided by the authors of these methods were used to conduct experiments. Note that, similar to \cite{2018-EnhancingtheSpatialResolutionofStereoImagesUsingaParallaxPrior-Jeon--,2018-FastAccurateAndLightweightSuperResolutionwithCascadingResidualNetwork-Ahn--}, EDSR \cite{2017-EnhancedDeepResidualNetworksforSingleImageSuperResolution-Lim--}, RDN \cite{2018-ResidualDenseNetworkforImageSuperResolution-Zhang--} and D-DBPN \cite{2018-DeepBackProjectionNetworksforSuperResolution-Haris--} are not included in our comparison since their model sizes are larger than our PASSRnet by at least 8 times.

\noindent
\textbf{Quantitative Results}
The quantitative results are shown in Table \ref{tab5}. It can be observed that our PASSRnet achieves the best performance on the Middlebury, KITTI 2012 and KITTI 2015 datasets. Specifically, compared to single image SR methods, our PASSRnet outperforms the second best approach (\emph{i.e.}, DRRN) by 1.04 dB in terms of PSNR on the Middlebury dataset for $2\times$ SR. Moreover, the PSNR value achieved by our network is higher than that of StereoSR by 1.00 dB. That is because, more reliable correspondence can be captured by our parallax-attention mechanism. 

\noindent
\textbf{Qualitative Results}
Figure \ref{fig7} illustrates the qualitative results achieved on \textcolor{black}{two} scenarios. It can be observed from zoom-in regions that single image SR methods cannot recover reliable details. In contrast, our PASSRnet uses stereo correspondence to produce finer details with fewer artifacts, such as the railings and stripe in Fig. \ref{fig7}. Compared to StereoSR, our PASSRnet explicitly captures stereo correspondence for SR. Consequently, superior visual performance is achieved.

\noindent
\textcolor{black}{
	\textbf{Flexibility}
	We further tested the flexibility of our PASSRnet and StereoSR \cite{2018-EnhancingtheSpatialResolutionofStereoImagesUsingaParallaxPrior-Jeon--} with respect to large disparity variations. Results achieved on images with different resolutions are shown in Table \ref{tab6}. More results under different baselines and depths are  available in the supplemental material. It can be observed that our PASSRnet is significantly better than StereoSR in terms of efficiency (\emph{i.e.}, FLOPs) on low resolution images. Meanwhile, our PASSRnet outperforms StereoSR by a large margin in terms of PSNR on high resolution images. That is because, StereoSR needs to perform padding for images with horizontal resolution lower than 64 pixels, which involves unnecessary calculations. For high resolution images, the fixed maximum disparity hinders StereoSR to capture longer-range correspondence. Therefore, the SR performance of StereoSR is limited.
}

\section{Conclusion}
In this paper, we propose a parallax-attention stereo super-resolution network (PASSRnet) to incorporate stereo correspondence for the SR task. Our PASSRnet introduces a parallax-attention mechanism with global receptive field to handle different stereo images with large disparity variations. We also introduce a new and the largest dataset for stereo image SR. It is demonstrated that our PASSRnet can effectively capture stereo correspondence for the improvement of SR performance. Comparison to recent single image SR and stereo image SR methods has shown that our network achieves the state-of-the-art performance.

{\small
	\bibliographystyle{unsrt}
	\bibliographystyle{ieee}
}

\end{document}